\pdfoutput=1
\documentclass[11pt]{article}
\usepackage{aaai2026}  
\usepackage{times}  
\usepackage{helvet}  
\usepackage{courier}  
\usepackage[hyphens]{url}  

\usepackage{graphicx} 
\usepackage{natbib}  
\usepackage{caption} 

\usepackage{latexsym}
\usepackage[utf8]{inputenc}
\usepackage{microtype}
\usepackage{inconsolata}
\newcommand{\abr}[1]{\textsc{#1}}
\usepackage{subfig}
\usepackage{xcolor}         
\usepackage{bm}
\usepackage{array}
\usepackage{multirow}
\usepackage{makecell}
\usepackage{xspace}
\usepackage{framed}
\usepackage{changepage}
\usepackage{cleveref}
\usepackage{enumitem}
\usepackage{siunitx}
\usepackage{tablefootnote}

\usepackage{booktabs}

\definecolor{formalshade}{rgb}{0.8, 0.9, 0.8}
\newenvironment{formal}{%
  \MakeFramed{\advance\hsize-\width\FrameRestore}%
  \noindent\hspace{-4.55pt}
  \begin{adjustwidth}{}{7pt}%
  \vspace{0.5pt}\vspace{0.5pt}%
}
{%
  \vspace{0.5pt}\end{adjustwidth}\endMakeFramed%
}

\title{PeerCoPilot: A Language Model-Powered Assistant for \\Behavioral Health Organizations}

\author{
  \textbf{Gao Mo\textsuperscript{*,1}},
  \textbf{Naveen Raman\textsuperscript{*,1}},
  \textbf{Megan Chai\textsuperscript{1}},
  \textbf{Cindy Peng\textsuperscript{1}},
  \textbf{Shannon Pagdon\textsuperscript{2}}, \\
  \textbf{Nev Jones\textsuperscript{2}},
  \textbf{Hong Shen\textsuperscript{1}},
  \textbf{Peggy Swarbrick\textsuperscript{3,}\textsuperscript{4}},
  \textbf{Fei Fang\textsuperscript{1}}}
\affiliations {
  \textsuperscript{1} Carnegie Mellon University, 
  \textsuperscript{2} University of Pittsburgh, \\
  \textsuperscript{3} Collaborative Support Programs of New Jersey
  \textsuperscript{4} Rutgers Graduate School of Applied and Professional Psychology 
}

\begin{document}
\maketitle
\begin{abstract}
Behavioral health conditions, which include mental health and substance use disorders, are the leading disease burden in the United States. 
Peer-run behavioral health organizations (\abr{pro}s) critically assist individuals facing these conditions by combining mental health services with assistance for needs such as income, employment, and housing. 
However, limited funds and staffing make it difficult for PROs to address all service user needs. 
To assist peer providers at \abr{pro}s with their day-to-day tasks, we introduce \textbf{\abr{PeerCoPilot}}, a large language model (LLM)-powered assistant that helps peer providers create wellness plans, construct step-by-step goals, and locate organizational resources to support these goals. 
\abr{PeerCoPilot} ensures information reliability through a retrieval-augmented generation pipeline backed by a large database of over 1,300 vetted resources. 
We conducted human evaluations with 15 peer providers and 6 service users and found that over 90\% of users supported using \abr{PeerCoPilot}. 
Moreover, we demonstrated that \abr{PeerCoPilot} provides more reliable and specific information than a baseline LLM. 
\abr{PeerCoPilot} is now used by a group of 5-10 peer providers at \abr{cspnj}, a large behavioral health organization serving over 10,000 service users, and we are actively expanding \abr{PeerCoPilot}'s use. 
\footnote{Our tool is available at \url{http://peercopilot.com} and code/dataset at \url{https://github.com/naveenr414/community_services_llm}}
\end{abstract}

\section{Introduction}
\label{sec:intro}
\begin{figure}[h]
    \centering 
    \includegraphics[width=0.45\textwidth]{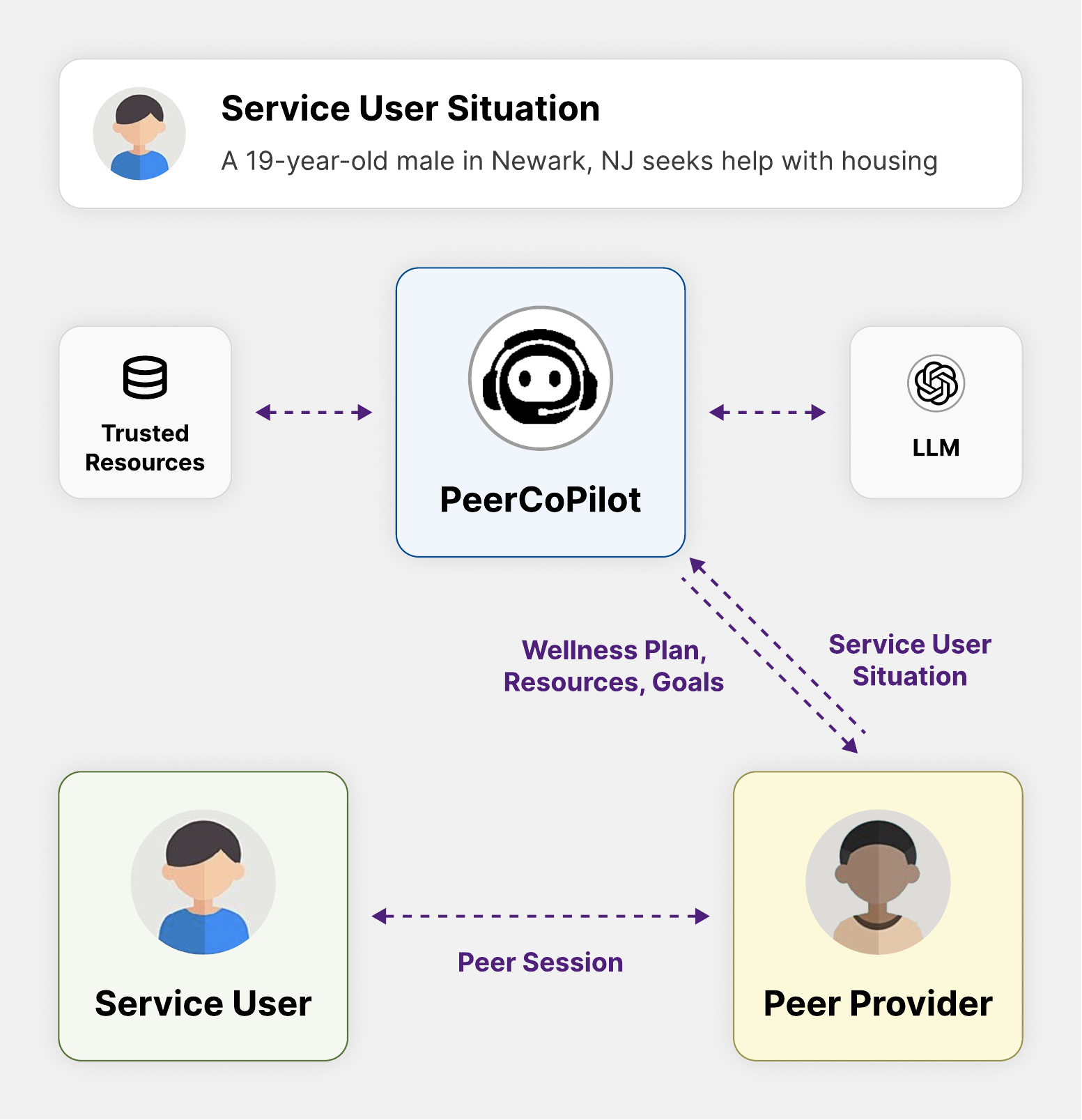}
    \caption{\abr{PeerCoPilot} assists peer providers at \abr{pro}s during peer sessions with service users. Because information reliability is critical, \abr{PeerCoPilot} combines LLMs with trusted sources to ensure accuracy.} 
    \label{fig:pull_small}
\end{figure}

Behavioral health conditions, including mental health and substance use disorders, are the leading disease burden in the United States, costing over \$80 billion annually~\cite{behavioral_health_cost}. 
Peer-run behavioral health organizations, referred to as \abr{pro}s, address this critical issue in difficult-to-engage communities facing disproportionately high rates of poverty, unemployment, and housing instability~\cite{schiz_cost,schiz_mortality}. 
\abr{pro}s tackle these issues through peer providers, who leverage their personal behavioral health experiences to provide service users with wellness support and resources for housing, financial, and employment resources~\cite{leadership_mental_health}.
\abr{pro}s are transformative for individuals with behavioral conditions; for example, in 2024, the Collaborative Support Programs of New Jersey (\abr{cspnj}) serves between 15 and 100 individuals daily through its network of 15 community wellness centers, reaches approximately 675 individuals through supportive housing programs, and supports around 700 individuals via wellness respites.

While service user demands grow year-to-year, \abr{pro} capacity has not kept up, leading to overburdened peer providers~\cite{wall2022experiences}. 
Increases in the prevalence of substance use and mental health disorders have led to growing service user demands~\cite{policy_behavioral_health}. 
At the same time, many \abr{pro}s are underfunded, limiting their ability to train and hire new peer providers, which becomes especially pressing due to high burnout rates for peer providers~\cite{capacity_peer_run}.
Adding to this burden is an underdeveloped technological infrastructure, with many \abr{pro}s lacking systems even to manage tracking the type and range of services and wellness supports they provide to people they serve.

To support \abr{pro}s, we propose using large language models (LLMs) as an assistant to peer providers.
LLMs have had success in other domains as a tool for information retrieval~\cite{tutor_copilot,llm_code,survey_copilot}. 
As a result, LLMs present a promising opportunity for \abr{pro}s by potentially helping peer providers craft tailored wellness plans and synthesize location-specific resources.
By assisting peer providers, LLMs can increase \abr{pro} capacity and service user capacity.
At the same time, while LLM-based assistants are prevalent~\cite{survey_copilot}, LLMs are rarely, if ever, used in \abr{pro}s because Many peer providers and service users have little familiarity with LLMs. 
These challenges necessitate a human-centered development process when introducing LLMs. 

In this paper, we introduce \textbf{\abr{PeerCoPilot}}, an LLM-based tool that assists peer providers with crafting wellness plans and retrieving resources. 
We developed \abr{PeerCoPilot} in partnership with \abr{cspnj}, a leading \abr{pro} based in New Jersey.  
After conversations with \abr{cspnj}, we designed \abr{PeerCoPilot} to assist with common tasks faced by peer providers, such as wellness plan creation, goal construction, resource recommendation, and benefit navigation. 
In our discussions with \abr{cspnj}, peer providers also stressed the need for reliable information when working with LLMs, so \abr{PeerCoPilot} ensures information reliability by combining an LLM-based backend with trusted resources through techniques such as retrieval augmented generation (\abr{rag})  (see Figure~\ref{fig:pull_small}).
We evaluated \abr{PeerCoPilot} through 3 onsite demos, 2 annotation sessions, and 1 semi-structured interview, totaling 15 peer providers and 6 service users. 
Through our onsite demos, we showed that peer providers and service users are willing to use \abr{PeerCoPilot}, and through our annotation sessions, we found that \abr{PeerCoPilot} provides more reliable and specific information than a baseline LLM. 
Our work is deployed to a group of peer providers at \abr{cspnj} who use \abr{PeerCoPilot} in their daily operations, and two other \abr{pro}s have reached out as a result of our initial deployment. 
\section{Related Works}
\paragraph{LLMs to Support Behavioral Health Professionals}
While there has been little work on LLMs in a behavioral health context, LLMs have seen great success in a variety of related fields, including education~\cite{cs_50,generative_ai_test,llm_scoring}, social science~\cite{agent_sense,ai_values,can_llm_social_science}, and mental health~\cite{lai2023psy,liu2023chatcounselor,beredo2022hybrid,crasto2021carebot}. 
Within behavioral health, most related are work in mental health on simulating patients~\cite{patient_psi,roleplay_doh} and work on using AI to answer substance use questions~\cite{substance_use_ai}. 
Unlike traditional mental health applications, which are often clinically focused, \abr{pro}s emphasize holistic wellness through housing, employment, and financial stability alongside behavioral health.

\paragraph{Copilot Tools} LLM-based copilot tools are used in domains such as software~\cite{ai_copilot_1,ai_copilot_2}, retail~\cite{copilot_retail}, and health~\cite{copilot_healthcare}. 
Copilot tools improve productivity by providing templates that scaffold development~\cite{impact_copilot}. 
However, copilot tools could reduce critical thinking skills and induce dependence~\cite{cognitive_effort_copilot}. 
In light of this, we develop \abr{PeerCoPilot} as a way to provide peer providers with extra resources, thereby augmenting rather than replacing them. 
\section{Background}
\label{sec:background}
To assist peer providers at \abr{pro}s, we developed \abr{PeerCoPilot}, an LLM-based tool for assisting peer providers during sessions with service users. 
We developed \abr{PeerCoPilot} based on conversations with peer providers and service users at \abr{cspnj}, as both groups value factually reliable and specific information. 
Reliability is key because peer providers work in settings where incorrect information can negatively impact service users.
For example, LLM-based tools could provide inaccurate information on benefit eligibility for government programs such as Medicare, which could mislead service users. 
Meanwhile, specific information allows peer providers to provide tailored information to service users. 
For example, more specific information could allow peer providers to generate a detailed step-by-step plan.

\begin{figure*}[h]
    \centering 
    \includegraphics[width=\textwidth]{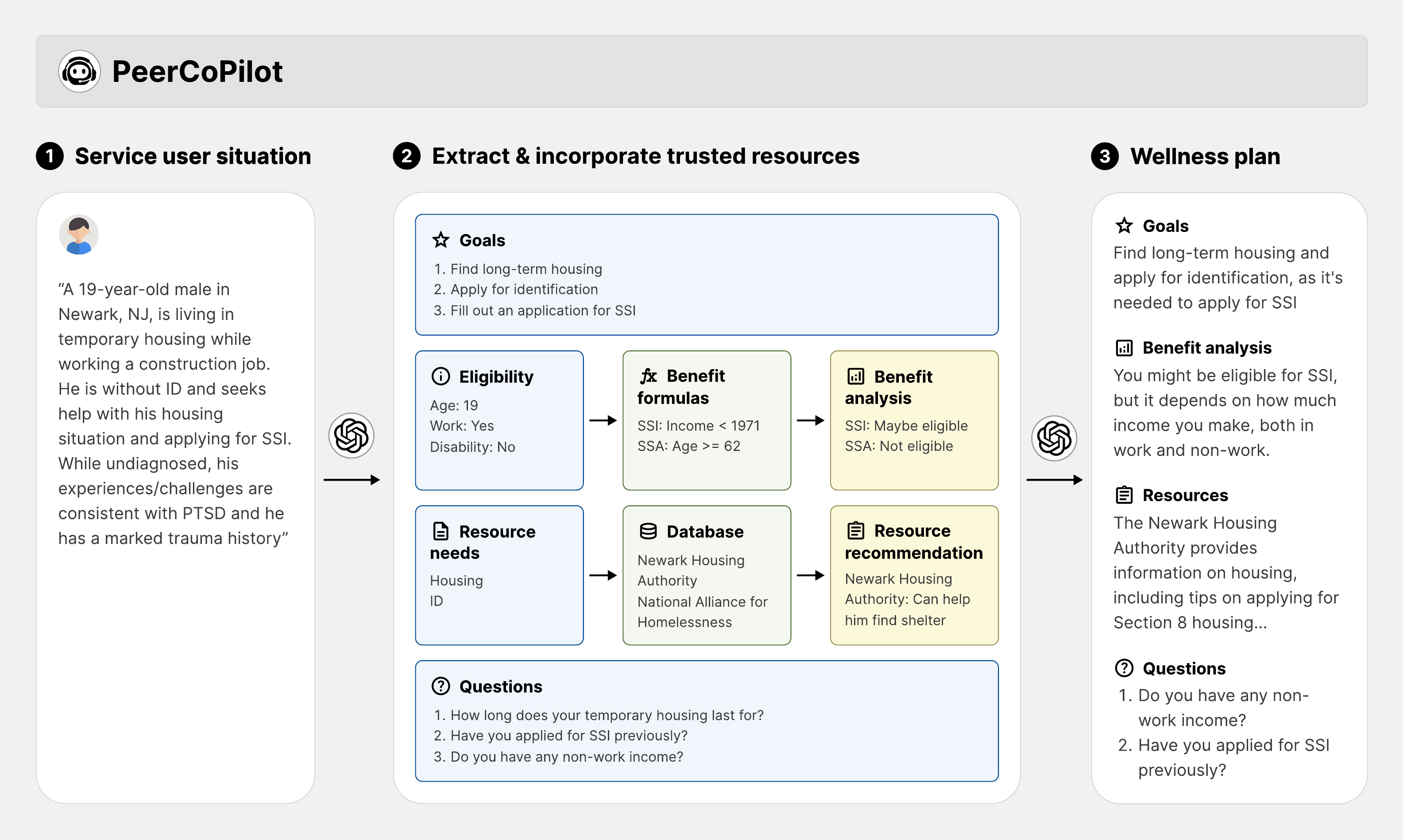}
    \caption{\abr{PeerCoPilot} takes in a peer provider's input and passes this into four modules:  resource recommendation, benefit eligibility, goal construction, and question generation. Some modules combine the peer provider's query with externally verified information to ensure accuracy. The results from all modules are combined to form a response that tackles various aspects, including goals, resources, and follow-up questions. } 
    \label{fig:pull}
\end{figure*}

Based on discussions with peer providers and service users, we constructed \abr{PeerCoPilot} as a chat-based assistant. 
\abr{PeerCoPilot} provides suggestions based on a service user's situation and answers peer provider questions.  
We ensured information reliability through modules such as resource recommendation and benefit analysis that combine trusted resources with an LLM (see Figure~\ref{fig:pull}). 
\section{System Design}
\label{sec:system_design}

\abr{PeerCoPilot} combines an LLM backend with modules that rely on verified information sources to ensure reliability. 
We describe the overall frontend and backend before describing individual modules.

\begin{figure*}[h]
    \centering 
    \includegraphics[width=\textwidth]{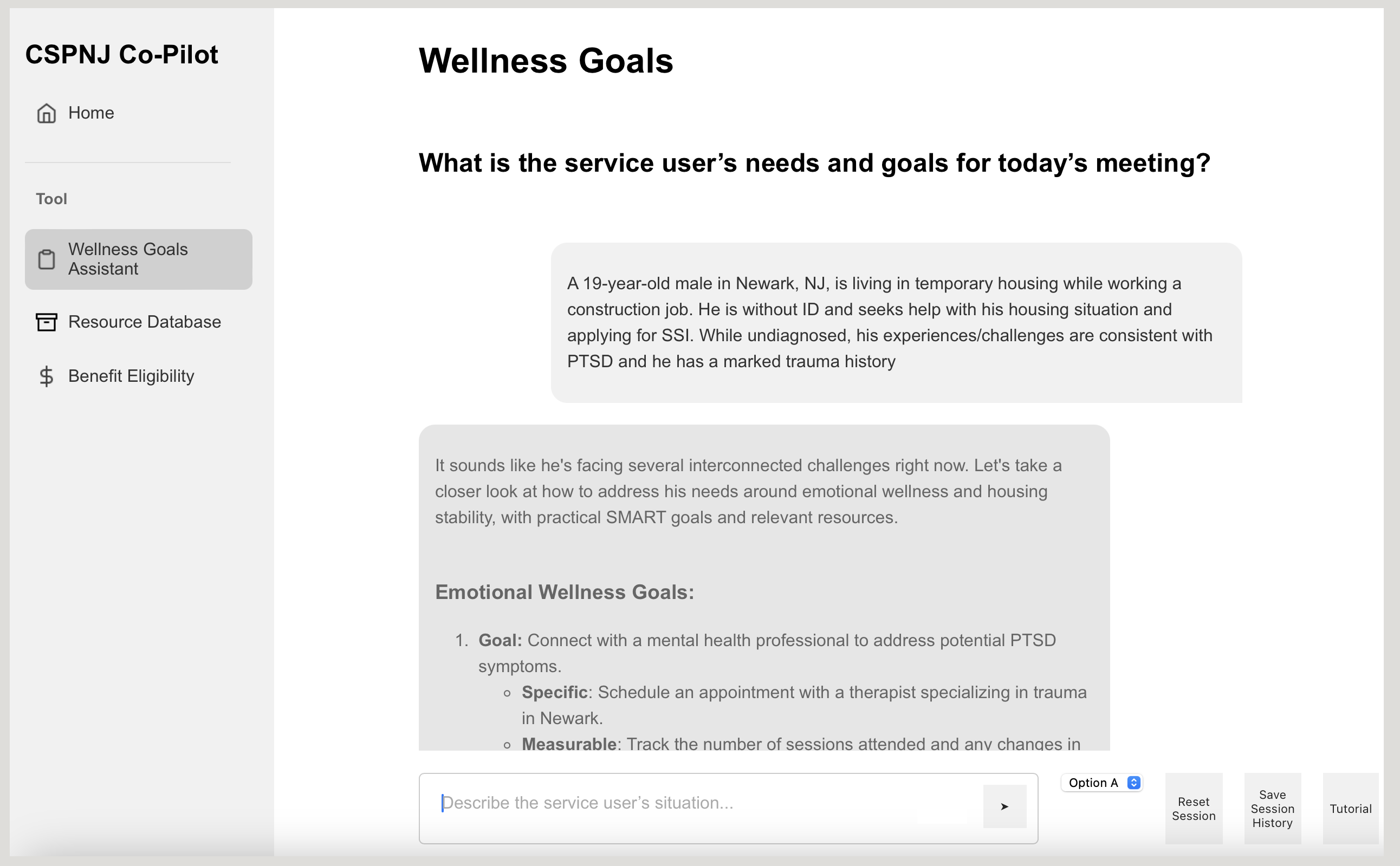}
    \caption{\abr{PeerCoPilot} is a chat-based frontend that presents peer providers with goals, resources, and follow-up questions pertinent to a service user's situation. Peer providers can also ask further questions to \abr{PeerCoPilot}.}
    \label{fig:ui}
\end{figure*}

\subsection{Frontend}
\abr{PeerCoPilot} consists of a chat-based frontend with buttons to reset and save the current session, along with a tutorial (see Figure~\ref{fig:ui} for a screenshot). 
The reset session button allows peer providers to clear the current session between peer sessions, while the save session history button allows for a written record if peer providers want to reference a session later. 
The tutorial button plays a three-minute video on how to use \abr{PeerCoPilot}. 
We built the frontend using \textsc{React} and \textsc{Socket.io}. 

\subsection{Backend Structure}
\abr{PeerCoPilot} crafts a response by aggregating information from modules. 
\abr{PeerCoPilot} relies on four modules: resource recommendation, benefit eligibility, goal construction, and question generation. 
After receiving outputs from all modules, \abr{PeerCoPilot} queries GPT-4 to craft a response using the information from each module. 
We instruct \abr{PeerCoPilot} to construct a response that holistically addresses the service user's situation following the eight dimensions of wellness framework (physical, spiritual, social, intellectual, financial, environmental, occupational, and emotional) used at \abr{cspnj} to guide peer providers~\cite{eight_dimensions}.

\subsection{Backend Modules}
\subsubsection{Resource Recommendation}
\label{sec:resources}
The resource recommendation module combines 
a resource database with \abr{rag} to ensure information reliability. 
Our database has over 1300 resources vetted by peer providers at \abr{cspnj}. 
Given a service user's background and goals, we use GPT-4 to extract resource needs, then match these with resource descriptions in the database via \abr{rag}~\cite{lewis2020retrieval}. \abr{rag} matches embeddings for resource needs with database entries. 
We construct embeddings using a SentenceTransformer with the MPNet v2 model~\cite{mpnet}, and retrieve according to the L2 metric.  

\subsubsection{Benefit Navigation} 
\label{sec:benefit}
Government benefits, such as Supplementary Social Income (SSI) and Medicaid, involve complex inclusion criteria, making it difficult to determine eligibility. 
Relying on GPT-4 for benefit eligibility can result in outdated or incorrect information. 
Instead, in the benefit eligibility module, we first use GPT-4 to extract demographic information such as age, monthly income, and total savings. 
We then pass this to formulas that assess eligibility given demographic information. 
These formulas are manually translated from eligibility information on government websites (e.g.,~\citet{ssi_eligibility}), and can be updated as requirements change.
This results in an assessment of whether a service user is likely to be eligible for each benefit.

\subsubsection{Goal Construction \& Question Generation}
\label{sec:wellness}
Peer providers need to offer support and construct plans tailored to service user situations. 
To assist with this, the goal construction module presents immediate goals for the service user, broken down into actionable steps. 
We construct goals by prompting GPT-4 with the SMART (Specific, Measurable, Achievable, Realistic, and Timely) goals framework~\cite{smart_goal}, as recommended by our partners at \abr{cspnj}. 
The question generation module suggests follow-up questions for peer providers to ask service users. 
It does so by prompting  GPT-4 to craft follow-up questions and additionally prompts with information on the dimensions of wellness~\cite{eight_dimensions} to ensure follow-up questions are holistic.
Examples include ``Do you have a stable place to stay?'' and ``Do you have transportation?''

\section{\abr{PeerCoPilot} Evaluation}
\label{sec:eval}
We evaluated \abr{PeerCoPilot} through human studies with peer providers and service users and found that both would use \abr{PeerCoPilot} in peer support sessions. 
We also found that \abr{PeerCoPilot} delivers more reliable and specific information than a GPT-4 baseline. 

\subsection{On-site Human Evaluations}
\begin{figure}[t]
    \centering 
    \includegraphics[width=0.5\textwidth]{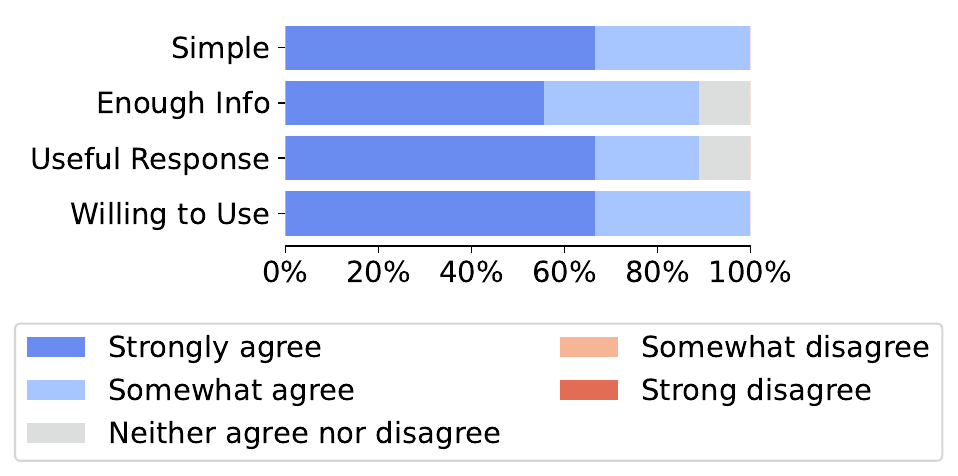}
    
    \caption{We surveyed peer providers on the usability of \abr{PeerCoPilot}. All peer providers found \abr{PeerCoPilot} simple and are willing to use it in practice.}
    \label{fig:usability}
\end{figure}

\begin{figure*}[h]
    \centering 
    \includegraphics[width=0.86\textwidth]{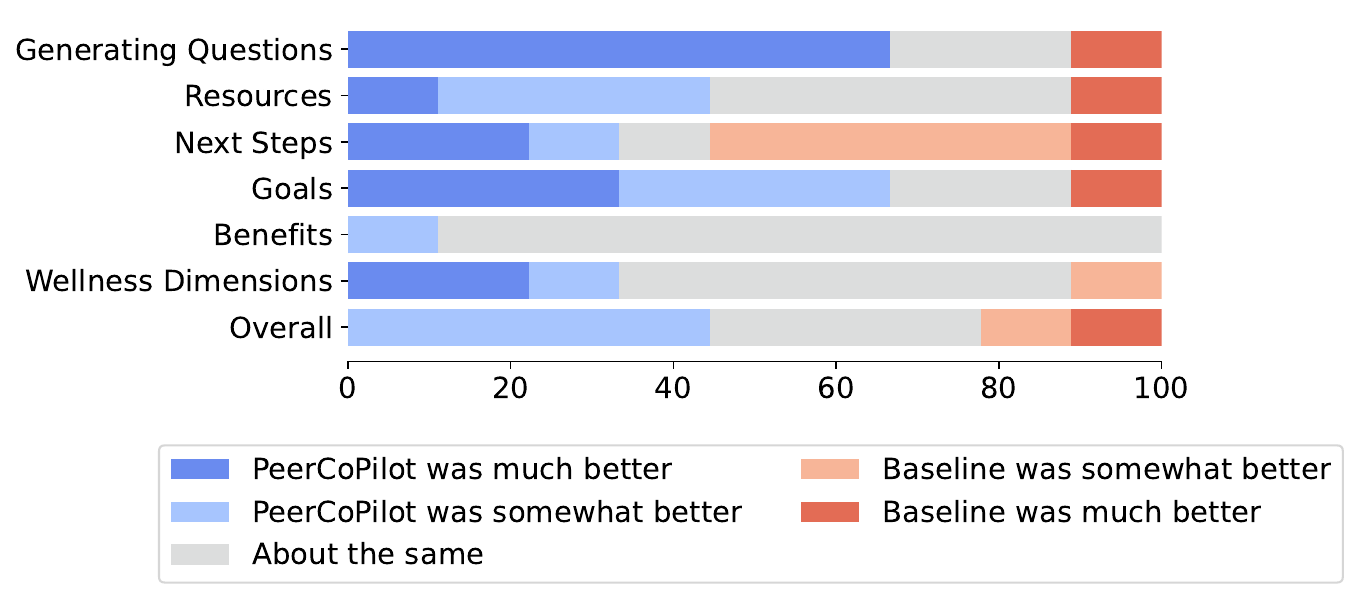}
    \caption{Peer providers found that \abr{PeerCoPilot} generates better questions, recommends better resources, and crafts better goals compared to a baseline. This is because the modules ensure information reliability and specificity.}
    \label{fig:comparison}
\end{figure*}

To assess \abr{PeerCoPilot}, we conducted an on-site study with nine peer providers and six service users. 
Participants interacted with \abr{PeerCoPilot} and baseline GPT-4 in random order to explore scenarios.
We constructed nine diverse scenarios capturing different types of situations faced by service users in reality; we further detail these scenarios in Appendix~\ref{sec:human_details}, and provide an example below: 
\begin{formal}
    A 19-year-old is in New Brunswick, NJ in living in temporary housing while working a construction job. He is without ID and seeks help with stabilizing his housing situation and improving financial wellness. While undiagnosed, his experiences/challenges are consistent with PTSD, and he has a marked trauma history.
\end{formal}
Participants then completed two surveys: one for system usability and another to compare \abr{PeerCoPilot} and the baseline (details in Appendix~\ref{sec:human_details}).
We include service user results in Appendix~\ref{sec:service_user}.
While we focus on results comparing against baseline GPT-4o mini, in Appendix~\ref{sec:automatic}, we detail our comparisons against other LLMs through the LLM-as-judge framework.

\paragraph{\textbf{Peer providers are willing to use \abr{PeerCoPilot}}}
In Figure~\ref{fig:usability}, we found all peer providers are willing to use \abr{PeerCoPilot} in practice. 
Peer providers found \abr{PeerCoPilot} simple to use, and 8 out of 9 peer providers believe that \abr{PeerCoPilot} delivered useful information for their queries. 
One peer provider remarked ``\textit{how we can develop a realistic plan. I love that...how it's breaking it down by dimension.}''
Another peer provider said how \abr{PeerCoPilot} can assist peer providers: ``\textit{I found that PeerCoPilot's follow-up questions and prompts would be crucial for a service provider to continue to assist someone in creating their wellness plan.}'' 

\begin{figure*}[h]
    \centering 
    \includegraphics[width=\textwidth]{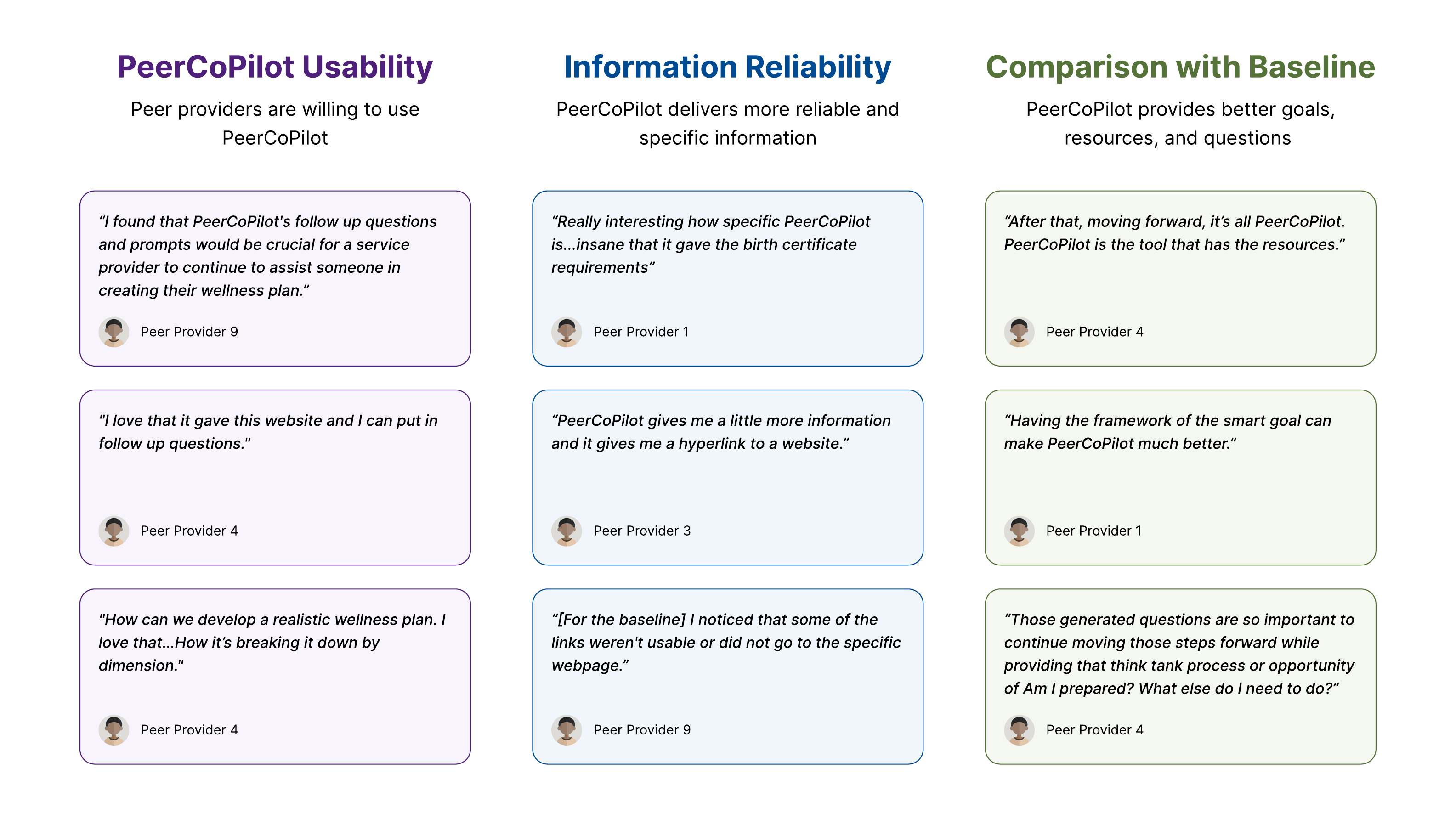}
    \caption{We outline three themes from our sessions with peer providers: 1) \abr{PeerCoPilot} provides useful information and peer providers are willing to use it, 2) \abr{PeerCoPilot} provides reliable and specific information, and 3) \abr{PeerCoPilot} provides better goals and questions than the baseline.}
    \label{fig:quotes}

\end{figure*}

\paragraph{\textbf{\abr{PeerCoPilot} delivers more reliable and specific resources}}
In Figure~\ref{fig:comparison}, we show that 4 out of 9 peer providers believe \abr{PeerCoPilot} delivered better resources, while only 1 out of 9 believed the baseline does. 
\abr{PeerCoPilot} delivers more reliable and specific information because it builds on top of a trusted database.
Peer providers notice this difference, as one remarks ``\textit{PeerCoPilot gives me a little more information and gives me a hyperlink to a website.}''
Peer providers also found \abr{PeerCoPilot} specific, with one noting that it was ``\textit{really interesting how specific PeerCoPilot is...insane that it gave the birth certificate requirements.}''
Conversely, for the baseline, one peer provider stated that they ``\textit{noticed that some of the links weren't usable or did not go to the specific webpage.}''

\paragraph{\textbf{6 out of 9 peer providers preferred \abr{PeerCoPilot} for goal construction and question generation (Figure~\ref{fig:comparison})}}
Peer providers found \abr{PeerCoPilot}'s SMART goals framework useful; one peer provider remarks ``\textit{Having the framework of the SMART goal can make PeerCoPilot much better}'' because it ``\textit{spells [the goal] out}.''
Peer providers also liked \abr{PeerCoPilot}'s follow-up questions: ``\textit{Those generated questions are so important to continue moving those steps forward while providing that think tank process or opportunity of Am I prepared? What else do I need to do?}'' 

\subsection{Reliability and Specificity}
\label{sec:eval_resource}
To understand whether \abr{PeerCoPilot} recommends more reliable and specific resources, we conducted an annotation study comparing resources from \abr{PeerCoPilot} and the baseline. 
We annotated resources according to whether correct contact information is present and resource specificity.
Additionally, two experts annotated resources based on usefulness, based on whether peer providers would recommend the resource; details in Appendix~\ref{sec:resource_details}. 

\begin{table*}[h]
\centering 
\caption{\abr{PeerCoPilot} provides contact information more frequently and provides more specific resources. When the underlying database is well populated (scen. 1), \abr{PeerCoPilot} achieves high effectiveness scores.}
\begin{tabular}{@{}lrrrlll@{}}
\toprule
Option    & Contact Provided & Bad Link & Verified & Specificity & Scen. 1 & Scen 2. \\ \midrule
\abr{PeerCoPilot} & 100\%             & 0\%      & 92\%          & 4.5/5       & 4.5/5   & 3.7/5   \\
Baseline  & 56\%              & 11\%       & 48\%        & 3.4/5       & 4.1/5   & 4.4/5   \\ \bottomrule
\end{tabular}
\label{tab:eval_resource}
\end{table*}

Our annotation results mirror the human evaluation, as \abr{PeerCoPilot} delivers more reliable and specific resources than the baseline (Table~\ref{tab:eval_resource}). 
\abr{PeerCoPilot} delivered resources that are 33\% more specific and 79\% more likely to provide contact info, while never giving inaccurate links. 
\abr{PeerCoPilot} identified more specific resources and more reliably provides correct contact information. 
Additionally, \abr{PeerCoPilot} delivered resources verified by \abr{cspnj} peer providers 92\% of the time, compared to 48\% for the baseline. 
When comparing the quality of resources across the two scenarios, we found that \abr{PeerCoPilot} delivered higher quality resources for scenario 1 (which focuses on health) while the baseline delivered higher quality resources for scenario 2 (which focuses on housing). 
\abr{PeerCoPilot} performed better in scenario 1 because the underlying database is better populated for health-related resources than housing-related ones. 
We compared the resources generated by each tool for two scenarios: the first scenario focuses on physical health, while the second scenario focuses on housing. 
This discrepancy underscores the benefits and drawbacks of relying on a verified database; when the database is well-populated (such as scenario 1), \abr{PeerCoPilot} delivers effective resources, while sparsely-populated databases (such as scenario 2) lead to poor performance. 
Expanding the underlying database can help ensure comprehensive verified resources. 

\subsection{Preliminary study on Peer provider-\abr{PeerCoPilot} Teaming}
\label{sec:human_ai_eval}
We conducted a pilot human-AI teaming study to assess \abr{PeerCoPilot}'s ability to help peer providers create wellness plans. 
We assessed the completion time and quality for wellness plans completed with and without \abr{PeerCoPilot}.
We show \abr{PeerCoPilot} reduced completion time by 10\% while leading to wellness plans better tailored to service user situations (details in Appendix~\ref{sec:human_ai_details}). 
Our study demonstrated the positive impact \abr{PeerCoPilot} can have when peer providers use it in practice. 
\section{Discussion and Path to Deployment}
\label{sec:discussion}
\abr{pro}s experience staffing shortages and low-tech solutions, inhibiting their ability to assist service users.
To tackle this, we present \abr{PeerCoPilot}, an LLM-based tool that assists peer providers at \abr{pro}s. 
\abr{PeerCoPilot} combines trusted resources with an LLM backend to ensure information reliability. 
Through human evaluations with 15 peer providers and 6 service users, we found that both groups would use \abr{PeerCoPilot}. 
We found that \abr{PeerCoPilot} provides more reliable and specific information than a baseline LLM. 
A group of peer providers at our partner \abr{pro}, \abr{cspnj}, use \abr{PeerCoPilot}. 

Our early results from the evaluation of \abr{PeerCoPilot} are promising, and already two additional \abr{pro}s have reached out with interest in deploying \abr{PeerCoPilot}.
Through our evaluation studies, we found that peer providers and service users are enthusiastic about \abr{PeerCoPilot}'s ability to assist \abr{pro}s. 
Possible improvements include a more user-friendly response; peer providers noted that some responses are verbose, making it difficult to quickly parse and understand. 
Additionally, we believe that features such as audio transcription and read-aloud (for low-literacy service users) and translation (for non-native English speakers) could improve adoption rates for our tool. 
Finally, expanding the resource database to be more comprehensive could improve the resource recommendation module. 

Beyond technical improvements, further evaluation studies are warranted. 
This includes an expansion of our human-AI teaming scenario from Section~\ref{sec:human_ai_eval} to a larger group of peer providers and service users, and a comparison against a larger set of baseline LLMs. 
We hope to accomplish these in the upcoming months and fully roll out our tool across \abr{cspnj}. 
Our early results are promising, and we are working towards expanding the deployment of \abr{PeerCoPilot}. 

\section*{Acknowledgements}
This work was supported by the National Science Foundation (NSF) program on Civic Innovation Challenge (CIVIC) under Award No. 2527408 and 2431230. 
We would like to thank everyone at CSPNJ for their feedback, comments, and collaboration throughout this project. 
We would additionally like to thank Sara McNemar and Christian Porter for their insights. 
We thank Stephanie Milani, Ananya Joshi, Gaurav Ghosal, and Jaskaran Walia for comments on an earlier draft of this paper. 
Co-author Raman is also supported by an NSF Fellowship. 

\bibliography{custom}

\appendix
\section*{Ethics Statement}
\label{sec:ethics}
We conduct all user studies under IRB with study number \textsc{study}\oldstylenums{2021\_00000481}. 
For each study, we first receive informed consent from participants through a consent form. 
Through this form, we verify that participants are 18 years or older. 
We recruited participants from \abr{cspnj}, our collaborating \abr{pro}. 
After each study, we pay participants \$60 per session. 
If a participant participates in multiple sessions, then we paid \$60 for each session. 
We stored all data in a private, secured location that is password protected. 
We stored de-identified information to maintain the privacy of participants. 
We additionally compensated all participants \$60 for each session and sessions last around an hour. 
For all sessions, we received informed consent from participants and have them fill out a consent and a demographic form. 
All data is stored privately, and we remove all personally identifiable information. 
We developed \abr{PeerCoPilot} in cooperation with peer providers and service users at \abr{cspnj} to augment peer provider capabilities rather than replace them. 

\section{Human Study Details}
\label{sec:human_details}
During our evaluation in Section~\ref{sec:eval}, we have participants interact with either PeerCoPilot or GPT-4 for ten minutes. 
The baseline is GPT-4o mini instructed with the following prompt: \textit{You are a Co-Pilot tool for \abr{cspnj}, a peer-peer mental health organization. Please provide helpful responses to the client}. 
PeerCoPilot uses GPT-4o mini whenever using a backend LLM. 
For each version, we kept the frontend the same, and blind participants to which tool they're interacting with by labeling them as `Option A' and `Option B.'
After each interaction, we had participants fill out a usability form, where we ask questions inspired by the system usability scale~\cite{system_usability}. 
In particular, we asked four question: 1) I found the tool simple to use, 2) I felt the tool gave enough information without being too much, 3) I think the tool delivers useful responses for my questions, 4) I would like to use this tool in my daily workflow. 
We had participants answer each question on a scale from strongly disagree to strongly agree. 
After interaction with both tools, we had participants compare their interaction with each along seven dimensions in Figure~\ref{fig:comparison}. 
1)  Proactively generating questions to ask service users 2) Providing resources that match the service user's needs 3) Suggesting next steps for the service user to meet immediate goals 4) Constructing actionable goals for service users 5) Providing comprehensive information on benefit systems (if applicable) 6) Holistically considering multiple dimensions of wellness and 7) Overall preference
For the session with service users, we had them compare the tools according to all criteria except the fist (question generation). 
We instructed participants to say aloud any thoughts they had during the study: 
\begin{formal}
Our goal is to evaluate an AI-based tool that assists peer specialists like you with supporting service users. For an overview of this study, we will first briefly go over the tool and give a quick demonstration. Second, we will present a scenario and have you interact with the tool as if you were working with a service user facing such a situation. We will have you interact with two different versions of the tool. Our goal is to understand whether such a tool is useful and which version works best for you, so pay attention to any differences between the two versions. After interacting with each version, we will ask a set of both structured and open-ended questions to get your feedback. But feel free to share your thoughts at any time during this study. 
\end{formal}

We constructed a set of scenarios for peer providers and service users to interact with. 
We conducted these scenarios in tandem with an expert in social work and \abr{pro}s. 
We constructed a draft scenario by initially instructing ChatGPT to construct a scenario by sampling values for the following: disability, substance use, gender, location (in New Jersey), age, government benefits, employment. 
We then manually edited these scenarios and discard scenarios that are too similar. 
We ended up with nine scenarios which tackle a variety of issues. 
We present one such scenario below: 
\begin{formal}
    A 19-year-old undocumented male immigrant in New Brunswick, NJ, is living in temporary housing while working a construction job. He is without ID and seeks help with stabilizing his housing situation, accessing legal resources for immigration support, and improving financial wellness. While undiagnosed, his experiences/challenges are consistent with PTSD and he has a marked trauma history.
\end{formal}
For sessions with peer providers, we assigned two different scenarios when working with the baseline and with PeerCoPilot. 
For service users, we gave them two scenarios for each tool, and let them select the scenario that best matches their situation. 

\begin{figure}
    \centering 
    \includegraphics[width=0.45\textwidth]{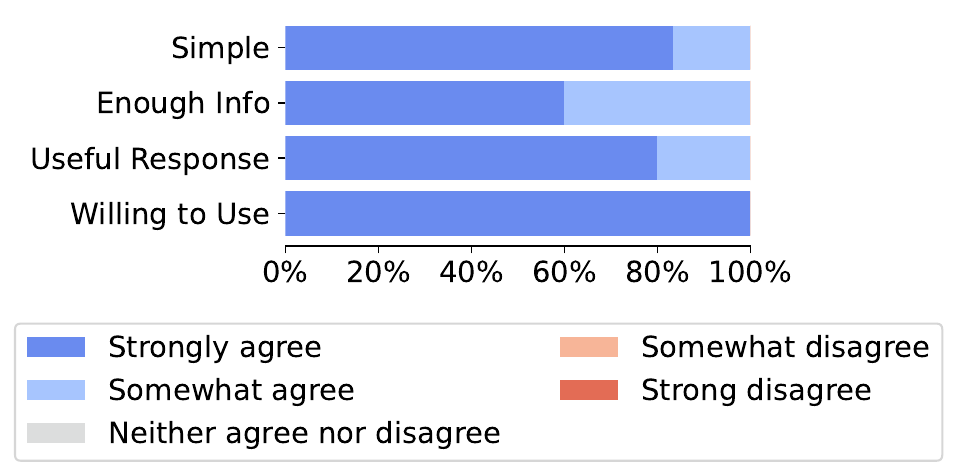}
    
    \caption{All service users found PeerCoPilot simple and providing useful responses. All service users support peer providers using PeerCoPilot in practice.}
    \label{fig:usability_service}
\end{figure}

\section{PeerCoPilot Service User Evaluation}
\label{sec:service_user}
For our sessions with service users, we found that all service are strongly in favor of peer providers using this tool (Figure~\ref{fig:usability_service}). 
Moreover, with service users, we found that all service users view our tool as easy to use and that it provides useful responses.
Taken with Figure~\ref{fig:usability}, we found that both service users and peer providers are heavily in favor of using our tool for peer sessions. 

\begin{figure}[h]
    \centering 
    \includegraphics[width=0.45\textwidth]{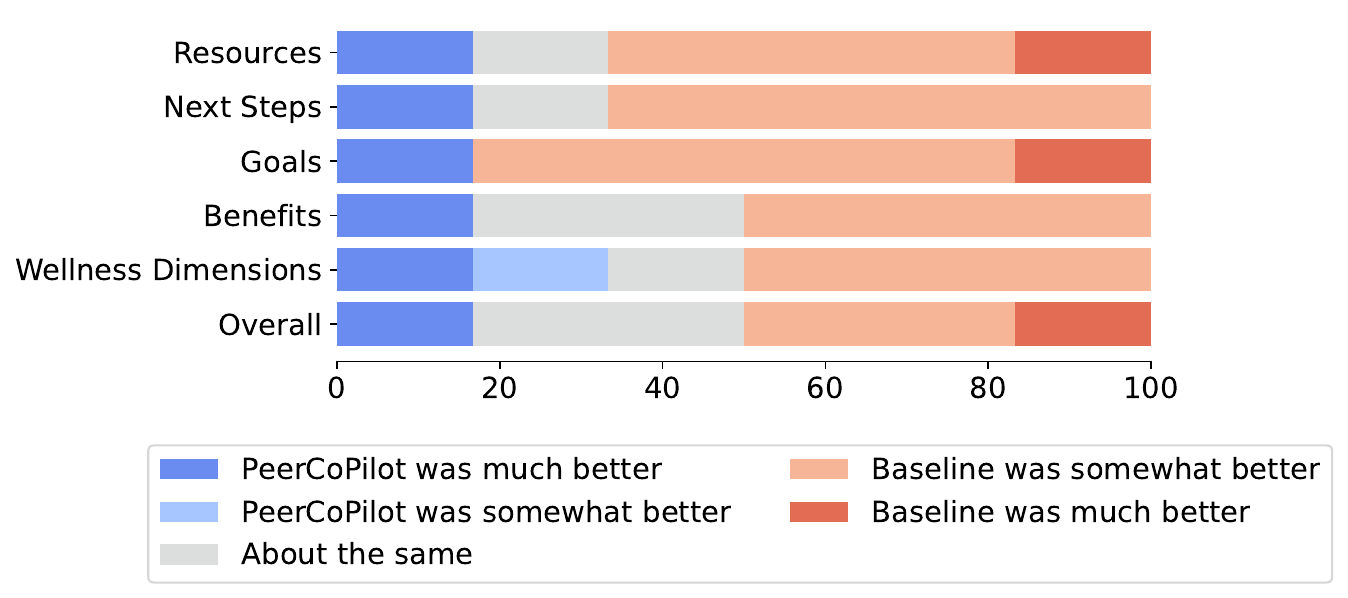}
    \caption{Service users prefer the baseline because of its simplicity, as it provides less information compared to PeerCoPilot. We note that some of these results were conducted using a translator, casting doubt on their results.}
    \label{fig:comparison_1b}
\end{figure}

Comparing between versions of the tool was more difficult for the service user evaluation due to language barriers which necessitated some user evaluations to be conducted with a translator.
We summarize our results in Figure~\ref{fig:comparison_1b}, and found that service users tend to prefer the GPT-4 baseline due to its simplicity. 
service users were generally unable to distinguish between the two tools or between the different criteria due to the language barriers. 
While service users enjoyed working with PeerCoPilot, we caution against generalizing the comparison results due to language difficulties. 

\section{Resource Evaluation Details}
\label{sec:resource_details}
We provide further details on the evaluation in Section~\ref{sec:eval_resource}. 
We generate resources for PeerCoPilot and the baseline by first querying the scenario and then providing an additional prompt to retrieve further resources: ``Can you provide specific resources for this scenario.''
We present two example scenarios below: 
\begin{formal}
    Scenario 1: A 38-year-old woman in Paterson, NJ is actively seeking physical therapy services to help her regain mobility and potentially return to full-time employment, but has limited knowledge about providers in her area. She has been living with her family for several months due to a physical disability that limits her ability to work full-time. She has a part-time job but cannot afford her medical expenses and is increasingly concerned about the sustainability of her current living situation. 
\end{formal}
\begin{formal}
    Scenario 2: A 60-year-old man in Newark is currently unhoused and staying in a temporary shelter after losing his job. He has a long history of alcohol use disorder and is in recovery, but he’s worried about his future housing stability. His main concern right now is finding permanent housing. He is struggling to find a place that will accept him due to his past, and he needs help connecting to local housing programs that can provide him with a long-term solution. Please provide resources for permanent housing.
\end{formal}
For annotation, we assessed according to the following criteria: 
\begin{enumerate}
    \item \textbf{Specificity} - Rate each resource on a 1-5 scale, where 5 refers to a resource that can be used directly, with a specific department or location mentioned for the resource at hand, while 1 refers to a resource that either does not exist or is a general purpose resource without being tailored towards the scenario at hand. 
    \item \textbf{Usefulness} - Rate each resource on a 1-5 scale, where 5 refers to a resource that an experienced peer provider would recommend for the scenario at hand, while 1 refers to a resource that no peer provider would recommend for the scenario at hand. 
    \item \textbf{Usability} - Rate each resource based on whether it provides \textbf{correct} contact details for each of the following modalities: a) address, b) phone number, and c) website. 
\end{enumerate}
We evaluated the specificity and usability using a single annotator, while we annotate the usefulness using expert peer provider annotators.

\section{Human-AI Team Evaluation Details}
\label{sec:human_ai_details}
We recruited three peer providers and had them construct wellness plans for four scenarios, two with the assistance of PeerCoPilot and two using other non-PeerCoPilot resources. 
For each pair, we measured the time to completion and quality of the wellness plan. 
We measured quality through a semi-structured interview with an expert peer provider, where we have the peer provider compare different wellness responses. 
We gave each peer provider at most 15 minutes to complete the wellness plan, and we instruct peer providers to complete wellness plans with 2-3 goals, 2-3 resources + next steps, and 2-3 follow-up questions. 

Without PeerCoPilot, peer providers took 10:20 to complete wellness plans, while with PeerCoPilot, we found that peer providers took 9:25. 
Because speed alone cannot dictate whether PeerCoPilot is useful, we assesed the quality of the wellness plans created in tandem with PeerCoPilot against those created separately. 
We found that PeerCoPilot can improve the quality of wellness plans. 
For example, during our semi-structured interview, the evaluator noted that the ``\textit{PeerCoPilot is more geared towards what scenarios is looking for.}''
In one scenario, the evaluator praised the combination of peer providers and PeerCoPilot for suggesting vocational rehabilitation (VR) and noted that ``\textit{lots of people don't know about it}'', and ``\textit{if PeerCoPilot brought it up, then that's good.}''
The evaluator consistently noted that the inclusion of PeerCoPilot improved the specificity of the wellness plan, independent of the peer provider who completed it. 
Taken together, through our human-AI teaming evaluation, we found that PeerCoPilot can allow peer providers to complete wellness plans quicker and with higher quality. 

\section{Automatic Evaluation Details}
\label{sec:automatic}
\begin{figure}
    \centering 
    \includegraphics[width=0.48\textwidth]{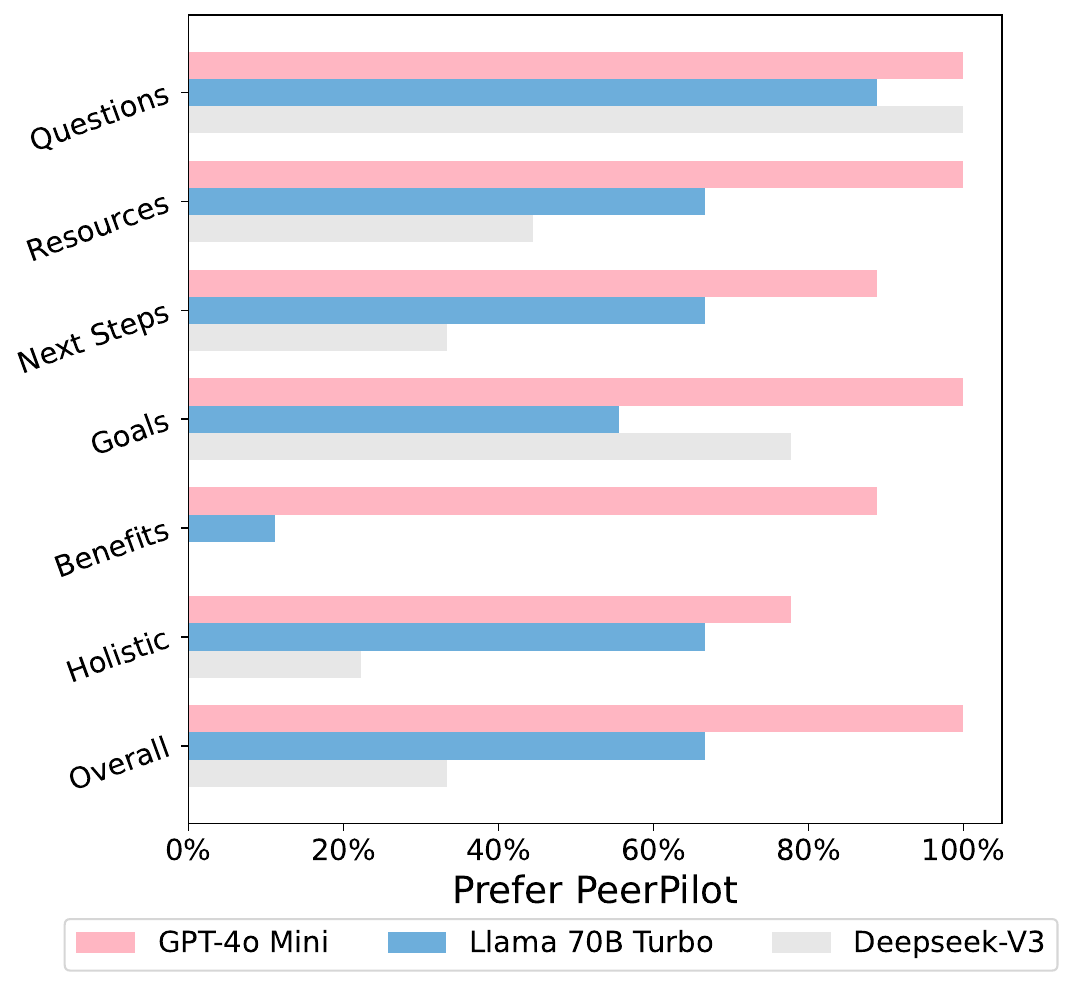}
    \caption{All LLM judges agreed that PeerCoPilot produces better follow-up questions and sets better goals than the baseline. Moreover, Llama and GPT view PeerCoPilot as finding better resources and being better overall.}
    \label{fig:gpt_comparison}

\end{figure}

To complement our human evaluations, we used the LLM-as-judge framework~\cite{llm_judge} to evaluate PeerCoPilot. 
We replicated the human evaluation study from Section~\ref{sec:eval} and have LLMs compare the output from PeerCoPilot and the baseline. 
We blind LLMs to which option is which, and we evaluate using the same nine scenarios from Section~\ref{sec:eval}. 
We used the following LLMs as judges: GPT-4o Mini~\cite{gpt_4}, Llama 70B turbo~\cite{llama}, and DeepSeek V3~\cite{deepseek_v3}, and compare them using the same criteria from Section~\ref{sec:eval}. 

In Figure~\ref{fig:gpt_comparison}, PeerCoPilot performed best at generating questions and setting goals across judges. 
Additionally, GPT and Llama found PeerCoPilot better overall, with improved performance compared to baselines in resource generation, next-step suggestions, and holistic wellness recommendations. 
While LLM-as-judge introduces an additional element of unreliability, we found that both human and LLM results consistently note that PeerCoPilot constructs better goals and generates better questions. 

\section{PeerCoPilot Prompt Details}
\label{sec:prompt_details}
We provide some of the prompts used for creating PeerCoPilot. 
PeerCoPilot stitches together modules through the following prompt: 

\texttt{
\small
You are a smart ChatBot that's associated with CSPNJ to help clients with their wellbeing. 
You will guide center service users along the different axes of wellness: emotional, physical, occupational, social, spiritual, intellectual, environmental, and financial
We will provide both a list of SmartGoals and potential resources, info on benefits, along with with a series of questions. 
The user will provide a situation, some Smart Goals, questions about the situation, and resources  
Please respond to the user using this information; you do not need to include all the information, just select what you think is most important. Only present information relevant to the user's situation. 
We need you to be concise yet thorough; you're chatting with the user, and you can always ask what they want more details on before providing the details 
Be thorough with the follow-up questions, and detail what situations this advice might work under
Pretend this is a normal chat with a user; don't present everything at once, but maybe one thing for this response (and provide others in later responses)
When presenting goals, align these explicitly along the dimensions of wellness 
When presenting resources, use only the resources that are provided by the user; don't try and make anything up, but use the things provided
Address everything in the third person; it's not the center service user who is asking these, but someone who is asking on behalf of them
You will be provided resources on some subset of transgender people, peer-to-peer support, crisis situations, and human trafficking/trauma. Please provide specific resources and outline SMART (Specific, Measurable, Achievable, Realistic, and Timely) goals in detail.
}

We constructed goals through the following prompt: 
\texttt{
You are a smart ChatBot that's associated with CSPNJ to help clients with their wellbeing. 
You will guide center service users along the different axes of wellness: emotional, physical, occupational, social, spiritual, intellectual, environmental, and financial
Provide SMART goals (Specific, Measurable, Achievable, Realistic, and Timely) tailored to the center service user’s needs. 
Try to be thorough}

Additionally, we construct follow-up questions through the following prompt: 
\texttt{
You are a smart ChatBot that's associated with CSPNJ to help clients with their wellbeing. 
You will guide center service users along the different axes of wellness: emotional, physical, occupational, social, spiritual, intellectual, environmental, and financial
Provide questions, such as details on their location and their situation, which can help better assist the center service user. Include explanations for why the question is important, and make sure you provide sufficient details about the questions and their explanations.}

The baseline operates through the following prompt: 
\texttt{
You are a Co-Pilot tool for CSPNJ, a peer-led mental health organization. Please provide helpful responses to the client.
}

\end{document}